\documentclass[journal]{IEEEtran}

\usepackage{comment}
\usepackage{epsfig}
\usepackage{graphicx}
\usepackage{amsmath}
\usepackage{amssymb}
\usepackage{comment}

\usepackage{multirow}

\usepackage{hyperref}
\hypersetup{hidelinks}

\usepackage{booktabs}
\usepackage{array, threeparttable}
\usepackage{caption}
\usepackage{subcaption}

\usepackage{algorithm}
\usepackage{listings}
\usepackage{colortbl}
\usepackage{color}
\usepackage[dvipsnames]{xcolor}
\usepackage{mathtools}
\usepackage{todonotes}
\usepackage{gensymb}
\usepackage{textcomp}

\usepackage{listings}
\definecolor{codegreen}{rgb}{0,0.6,0}
\definecolor{codegray}{rgb}{0.5,0.5,0.5}
\definecolor{codepurple}{rgb}{0.58,0,0.82}
\definecolor{backcolour}{rgb}{0.95,0.95,0.92}
\lstdefinestyle{mystyle}{
    backgroundcolor=\color{backcolour},
    commentstyle=\color{codegreen},
    keywordstyle=\color{magenta},
    numberstyle=\tiny\color{codegray},
    stringstyle=\color{codepurple},
    basicstyle=\ttfamily\footnotesize,
    breakatwhitespace=false,
    breaklines=true,
    captionpos=b,
    keepspaces=true,
    numbers=left,
    numbersep=5pt,
    showspaces=false,
    showstringspaces=false,
    showtabs=false,
    tabsize=2
}
\def\etal{\emph{et al}.}

\begin{document}

\title{A Simple yet Effective Network based on Vision Transformer for Camouflaged Object and Salient Object Detection}

\author{Chao Hao, 
Zitong Yu$^\dagger$,~\IEEEmembership{Senior Member,~IEEE}, 
Xin Liu$^\dagger$,~\IEEEmembership{Senior Member,~IEEE}, 
Jun Xu,~\IEEEmembership{Member,~IEEE}, 
Huanjing Yue,~\IEEEmembership{Senior Member,~IEEE} and
Jingyu Yang,~\IEEEmembership{Senior Member,~IEEE}




\thanks{$\dagger$Corresponding authors: Xin Liu (email: linuxsino@gmail.com) and Zitong Yu (email: yuzitong@gbu.edu.cn)}

\thanks{Chao Hao is with the School of Electrical and Information
Engineering, Tianjin University, Tianjin, 300072, China, and the School of Computing and Information Technology, Great Bay University, Dongguan, 523000, China (e-mail: 3018234336@tju.edu.cn).}

\thanks{Z. Yu is with the School of Computing and Information Technology, Great Bay University, Dongguan, 523000, China (e-mail: yuzitong@gbu.edu.cn).}

\thanks{Xin Liu is with the School of Electrical and Information
Engineering, Tianjin University, Tianjin, 300072, China, and the Computer Vision and Pattern Recognition Laboratory, Lappeenranta-Lahti University of Technology LUT, 53850 Lappeenranta, Finland (e-mail: linuxsino@gmail.com).}

\thanks{Jun Xu is with the School of Statistics and Data Science, Nankai University, Tianjin, 300072, China (e-mail: nankaimathxujun@gmail.com).}

\thanks{Huanjing Yue and Jingyu Yang are with the School of Electrical and Information Engineering, Tianjin University, Tianjin, 300072, China (e-mail: huanjing.yue@tju.edu.cn, yjy@tju.edu.cn).}

}

\markboth{IEEE TRANSACTIONS ON IMAGE PROCESSING}%
{Shell \MakeLowercase{\textit{et al.}}: Bare Demo of IEEEtran.cls for IEEE Journals}

\maketitle

\begin{abstract}
Camouflaged object detection (COD) and salient object detection (SOD) are two distinct yet closely-related computer vision tasks widely studied during the past decades. Though sharing the same purpose of segmenting an image into binary foreground and background regions, their distinction lies in the fact that COD focuses on concealed objects hidden in the image, while SOD concentrates on the most prominent objects in the image. Previous works achieved good performance by stacking various hand-designed modules and multi-scale features. However, these carefully-designed complex networks often performed well on one task but not on another. In this work, we propose a simple yet effective network (SENet) based on vision Transformer (ViT), by employing a simple design of an asymmetric ViT-based encoder-decoder structure, we yield competitive results on both tasks, exhibiting greater versatility than meticulously crafted ones. Furthermore, to enhance the Transformer's ability to model local information, which is important for pixel-level binary segmentation tasks, we propose a local information capture module (LICM). We also propose a dynamic weighted loss (DW loss) based on Binary Cross-Entropy (BCE) and Intersection over Union (IoU) loss, which guides the network to pay more attention to those smaller and more difficult-to-find target objects according to their size. Moreover, we explore the issue of joint training of SOD and COD, and propose a preliminary solution to the conflict in joint training, further improving the performance of SOD. Extensive experiments on multiple benchmark datasets demonstrate the effectiveness of our method. The code is available at  \href{https://github.com/linuxsino/SENet}{https://github.com/linuxsino/SENet}. 
\end{abstract}


\IEEEpeerreviewmaketitle

\section{Introduction}

\IEEEPARstart{C}{amouflaged} object detection (COD) aims to detect and segment objects ``seamlessly'' integrated into surrounding environments \cite{SINet, tip3} and salient object detection (SOD) aims to identify the most significant objects or regions in an image and segment them \cite{M3Net, tip1, tip2}. The essence of both tasks is to perform binary segmentation on the given image, dividing it into two parts: the target region and the non-target region \cite{binary-seg}, so they might be suitable for the same network architecture and training scheme (the training data used by the two tasks is different).

However, few previous works can perform well on both tasks at the same time \cite{hybrid, egnet, SINet, anabranch}. This can be explained by the fact that previous works were accustomed to manually designing various modules according to the characteristics of specific tasks \cite{ZoomNet, SegMAR, BBRF, basnet}, and achieving excellent performance on specific tasks through the network with complex architecture formed by stacking a large number of these modules \cite{EVP}. However, these specially designed modules also make the network only an expert in a specific field and lose its generality.

Based on the above viewpoint and inspired by the simple image-to-image architecture employed in MAE \cite{MAE} pre-training, we propose a simple yet effective network (SENet) based on vision Transformer (ViT) \cite{ViT} for COD and SOD. SENet comprises only a few fundamental modules, eliminating the need for the intricate stacking of numerous meticulously designed modules and avoiding the repetitive superimposition of multi-scale features, thus greatly simplifying the low-level structure segmentation models.

We find that image reconstruction is a beneficial auxiliary task \cite{auxiliary} for COD and SOD, which not only aids the network in extracting more low-level information essential for segmentation but also accelerates training. Consequently, we leverage reconstruction and segmentation together as training targets for SENet. As for why it is useful, a possible explanation is that masking the image and subsequently reconstructing it increases the overall training difficulty, compelling the network to enhance its perception of the entire image \cite{rppg}. To achieve this, an asymmetric encoder-decoder architecture is developed, with an encoder that operates only on the visible subset of patches (without mask tokens), along with a lightweight decoder that reconstructs the original image and generates the prediction map from the latent representation and mask tokens \cite{MAE}.

\begin{figure}[t]
      \centering
      \includegraphics[width=0.9\linewidth]{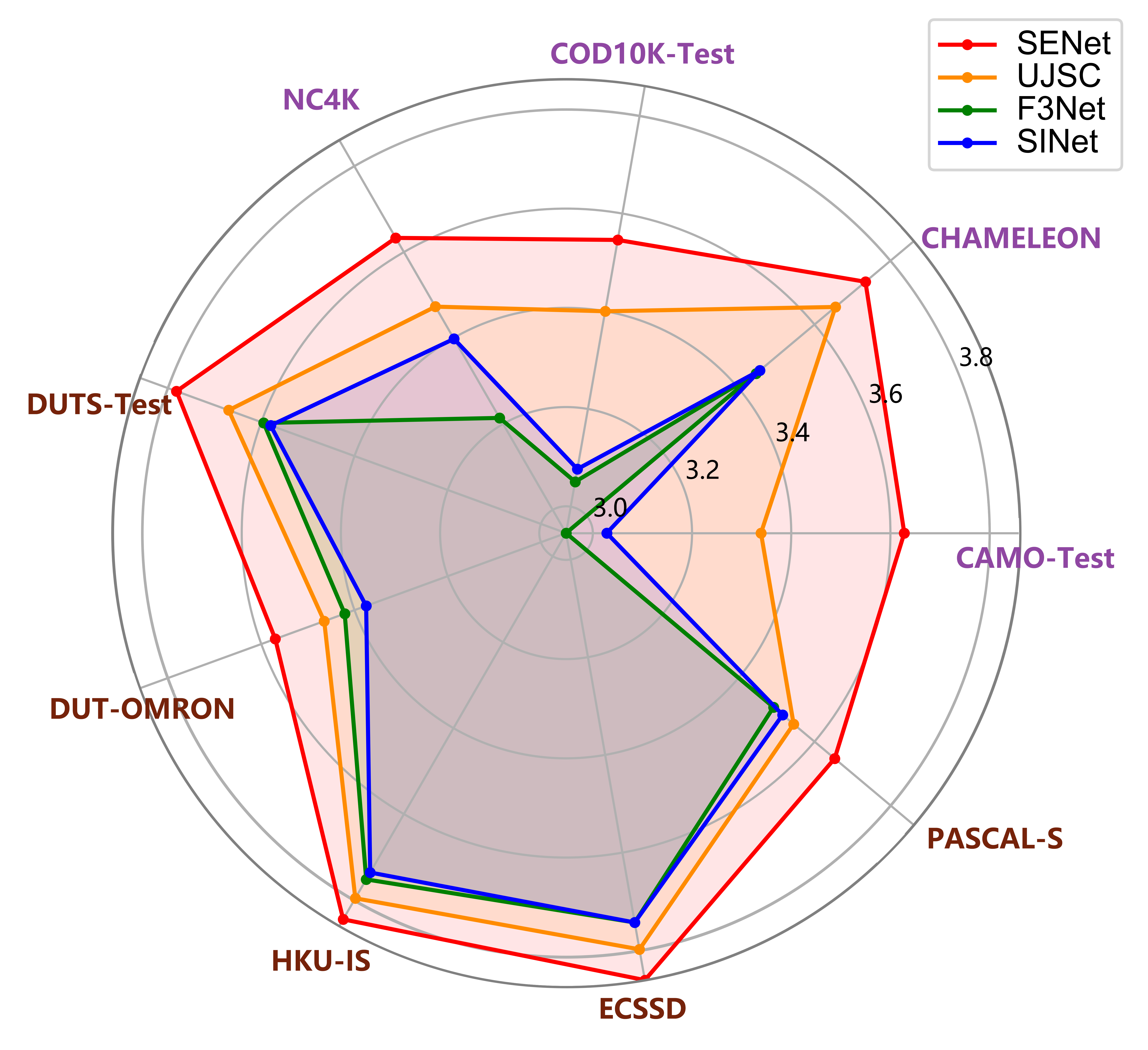}
      \caption{\small{SENet (Ours) achieves the best performance on nine datasets of COD (in \textcolor{Purple}{purple}) and SOD (in \textcolor{Brown}{brown}) compared with methods UJSC \cite{UJSC}, F3Net \cite{f3net} and SINet \cite{SINet}. The performance on the COD datasets is evaluated by testing the model trained on COD training sets, and the same applies to SOD. The more challenging the dataset, the greater the lead achieved by SENet. For the evaluation metric used here, see Eq.~(\ref{eq6}).}}
      \label{fig:radar}
   \end{figure}
   
SENet is mainly based on the self-attention mechanism, which is a patch-level modeling method for image \cite{Transformer, ViT, au}. To make it more suitable for the pixel-level COD and SOD tasks \cite{pixel-level}, we propose a local information capture module (LICM), a plug-and-play module for Transformer blocks. Specifically, LICM captures local information by using a small-kernel convolution within the neighboring pixels of a patch, and then fuses it with global information extracted by the Transformer block through residual linking \cite{ResNet, convpass}.  

Moreover, we propose a dynamic weighted loss (DW loss) based on Binary Cross-Entropy (BCE) and Intersection over Union (IoU) loss for a common problem in COD and SOD, that is, the smaller the size of the target object, the more challenging the segmentation becomes \cite{ZoomNet, f3net}. Specifically, we assign different weights to each region of the image according to the size of the target object, with smaller target objects being assigned higher weights. This allows the network to penalize prediction errors for small targets more significantly. In addition, the bilinear interpolation method is utilized to set new ground truth for pixels in the boundary area (transition between target and non-target areas) to deal with the labeling uncertainty \cite{labeling-uncertain, UJSC-V2} problem.

Our SENet has good performance on COD and SOD tasks, respectively. As shown in Fig.~\ref{fig:radar}, SENet achieves the highest scores on nine datasets for both tasks. In addition, we conduct experiments to explore its performance in cross-domain evaluation \footnote{\label{footnote:1} \scriptsize The cross-domain evaluation in this work refers to using the model trained on the COD datasets to test on the SOD datasets or using the model trained on the SOD datasets to test on the COD datasets.} and joint training (use the data of both to train the model together). We propose two joint training paradigms (see Fig.~\ref{fig:joint_training}) and Paradigm 2 preliminarily mitigates the problem of significant performance degradation in COD task during joint training due to the contradictory attributes \cite{UJSC, UJSC-V2} of the two tasks and further improves the performance on SOD task. 

In summary, our main contributions include:

\begin{itemize}

    \item  We propose the SENet, a simple yet effective network based on ViT, which is a general architecture and performs well on both COD and SOD. Moreover, we find that using image reconstruction as an auxiliary task can improve the performance of the network on both tasks.

    \vspace{1mm}
    \item  We propose the LICM, a simple plug-and-play module for Transformer blocks, which can enhance the Transformer's modeling ability of local information, making it more suitable for pixel-level COD and SOD tasks. 

    \vspace{1mm}
    \item  We propose the DW loss, which can actively adjust the weights according to the size of the target region when calculating the loss, further improving the network's ability to handle more difficult instances. 

    \vspace{1mm}
    \item We explore the issue of joint training of COD and SOD and discuss the impact of joint training on the performance of the two tasks, and propose a preliminary feasible solution to optimize joint training. We also report the results of joint training and cross-domain evaluation, demonstrating the contradiction between the two tasks.

    \vspace{1mm}
    \item We conduct experiments on nine benchmark datasets for COD and SOD and achieve state-of-the-art results.

\end{itemize}

\section{Related Work}
\label{sec:relatedwork}

\begin{figure*}[t]
      \centering
      \includegraphics[width=1.0\linewidth]{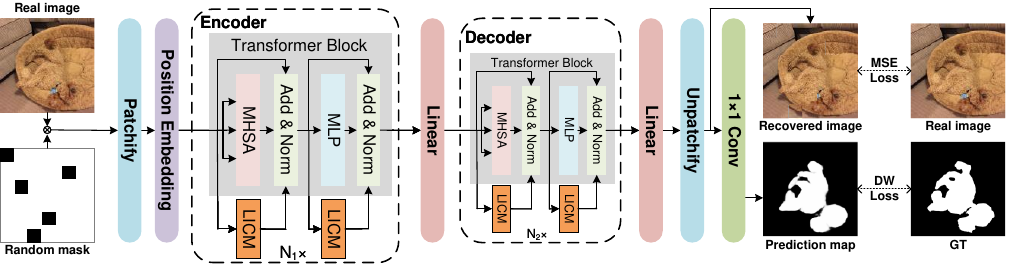}
      \caption{\small \textbf{Illustration of the proposed SENet with its training process}. Our SENet mainly consists of two parts: an encoder and a decoder composed of two asymmetric ViTs. The proposed LICM is seamlessly integrated in parallel with both the multi-head self-attention (MHSA) layer and the multi-layer perceptron (MLP) layer within every Transformer block. Masked images are utilized as input, and supervised training on the network is conducted by employing both the loss for the image reconstruction task and the loss for the binary segmentation task. $\otimes$
      indicates pixel-wise multiplication. See the Appendix for the illustration of the inference process of SENet.}
      \label{fig:SENet}
   \end{figure*}
   
\subsection{Salient and Camouflaged Object Detection}
\noindent \textbf{Salient object detection}. In the early days, researchers used hand-designed operators to extract features, the classic methods utilized low-level cues \cite{EVP-V2} such as color, brightness, and texture to detect salient objects \cite{SOD1, SOD2, SOD3, SOD4}. In recent years, deep learning-based \cite{f3net, LDF, SelfReformer} methods have shown better performance than traditional methods. Zhao \etal \cite{egnet} proposed an edge guidance network to better exploit the complementarity between salient edge information and salient object information.
Wu \etal \cite{CPD} designed a cascaded partial decoder to achieve accurate saliency detection with finer details. Wei \etal \cite{LDF} proposed a label decoupling framework which consists of a label decoupling procedure and a feature interaction network to solve the problem of edge pixels having a very unbalanced distribution. Li \etal \cite{li2023rethinking} designed a lightweight framework while maintaining satisfying competitive accuracy to achieve a favorable balance between efficiency and accuracy. Wang \etal \cite{tip7} designed a Curiosity-driven Network and a Curiosity-driven Learning Algorithm based on fragment attention mechanism newly defined to better solve the problem of computational redundancy.

\vspace{1mm}
\noindent \textbf{Camouflaged object detection}. Since the first large-scale and high-quality COD dataset was released \cite{SINet}, a lot of work \cite{BGNet, HitNet, SegMAR, coddiffusion} began to emerge, making the field of COD develop rapidly. Fan \etal \cite{SINet-V2} proposed a method of searching first and then confirming inspired by the hunting process of animals. Sun \etal \cite{BGNet} proposed a boundary-guided network to force the model to generate features that highlight object structure, thereby promoting camouflage detection of accurate boundary localization. Lv \etal \cite{NC4K} presented the first ranking-based COD network to simultaneously localize, segment and rank camouflaged objects, which better utilized priors on the degree of camouflage. Zhou \etal \cite{tip5} proposed a novel FAP-Net for the COD task, which can
effectively integrate cross-level features and propagate
the valuable context information from the encoder to the
decoder for accurately detecting camouflaged objects. Song \etal \cite{tip4}  introduced the concept of focus areas that represent some regions containing discernable colors or textures, and developed a two-stage focus scanning network for camouflaged object detection. Zhai \etal \cite{tip6} proposed a novel Mutual Graph Learning (MGL) model by shifting the conventional perspective of mutual learning from regular grids to graph domain.

These methods for the above two tasks are specially designed based on certain characteristics of COD or SOD, which are different from our starting point for designing SENet. SENet is a general architecture for both tasks, and only a general architecture that is suitable for both tasks has the potential to be applied to joint training.

\subsection{Masked Autoencoder}
\noindent Masked autoencoder (MAE) \cite{MAE} is a large-scale self-supervised pre-trained model, which achieved good performance on multiple downstream tasks \cite{MAE-down1, MAE-down2, MAE-down3, MAE-down4} and proved that image reconstruction is an effective self-supervised learning task for learning the underlying features of images. MAE is a network with a very elegant image-to-image structure, which uses an asymmetric encoder-decoder architecture consisting of two ViTs to complete image reconstruction, and has left a deep impression on us in reconstructing the image (even an image with a high masking ratio can be reconstructed well by MAE). Inspired by MAE, we propose the SENet, which is also an image-to-image network of generating a binary image from a RGB image, and we also use image reconstruction as one of the training objectives of SENet. MAE has been applied in many deep learning fields, but as far as we know, we are the first to directly use this elegant structure of MAE to directly complete the segmentation task.

\section{Methodology}
\label{sec:method}

\begin{figure}[t]
      \centering
      \includegraphics[width=1.0\linewidth]{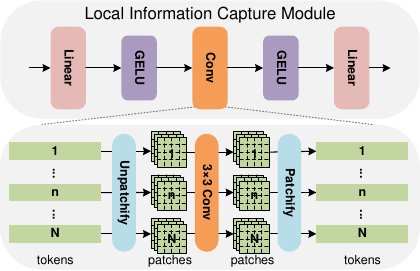}
      \caption{\small Illustration of the proposed LICM. ``Unpatchify'' and ``Patchify'' here are essentially reshape operations. LICM is able to capture more localized information by converting tokens into patches and then performing a small-kernel convolution.}
      \label{fig:licm}
   \end{figure}

\subsection{Overall Architecture}
\label{sec:architecture}

Fig.~\ref{fig:SENet} illustrates the overall training framework of the proposed SENet. SENet is a simple single-pathway model consisting only of some fundamental modules, generating single-channel prediction maps from input RGB images through an asymmetric ViT-based encoder-decoder structure \cite{MAE}. It can be seen that we use both image reconstruction and binary segmentation tasks as our training targets, where image reconstruction is the auxiliary task \cite{auxiliary, auxiliary2}. Specifically, the input image is first serialized into tokens where a small portion is randomly masked, then the encoder operates on the remaining tokens to model global and local contexts. After that, the decoder restores the removed tokens from the latent representation. Finally, the processed tokens are transformed back into the original image shape and the required single-channel prediction map is obtained through a $1 \times 1$ convolution operation. 

\subsection{Local Information Capture Module}
\label{sec:LICM}

The core attention mechanism of Transformer brings powerful global context modeling capabilities but lacks a locality mechanism for information exchange \cite{FSP} within a local region (e.g., a small block of 3$\times$3 pixel size). Besides, it is well known that COD and SOD are pixel-level segmentation tasks, which require the model to be able to capture more fine-grained information \cite{dense}. Inspired by \cite{restormer, convpass}, we propose the LICM, a plug-and-play module for Transformer blocks which is applied on a single patch to strengthen the local feature representation, making our ViT-based SENet more suitable for pixel-level COD and SOD tasks.

As illustrated in Fig.~\ref{fig:licm},  LICM consists of three learnable layers: two linear layers to change the dimension of the 1D tokens and a $3 \times 3$ convolution layer with the same input and output channel.  Before convolution, the 1D tokens are transformed into patches with 2D structure. Subsequently, a small-kernel convolution is applied to each individual patch to extract more localized information. Finally, 2D patches are reverted back to 1D tokens. We insert the LICM in parallel with the MHSA layer and MLP layer of each Transformer block \cite{Transformer, convpass}, which can be formulated as:

\begin{equation}
\label{eq1}  
\boldsymbol{X}'\leftarrow\boldsymbol{X}+\textit{LN}(\textit{MHSA/MLP}(\boldsymbol{X}))+\textit{LN}(\textit{LICM}(\boldsymbol{X})),
\end{equation}

where $\boldsymbol{X}'$ is the output and $\boldsymbol{X}$ is the input, \textit{LN} is Layer Normalization \cite{ln}.

\textbf{Relation to prior works}. There have been some ViT-based methods before that also proposed some modules for extracting local information \cite{FSP, convpass, dense}, but the main difference from our method is that we extract information within a smaller region (previous methods involve information exchange among adjacent patches, our SENet conducts information exchange within each individual patch).

\begin{figure}[t]
      \centering
      \includegraphics[width=1.0\linewidth]{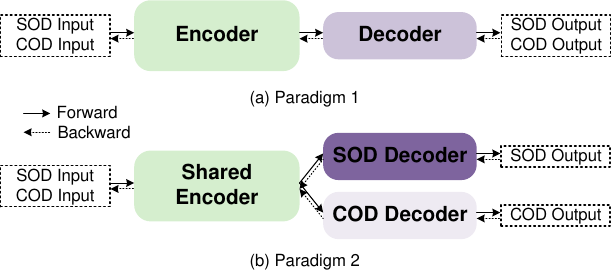}
      \caption{\small Illustration of two joint training paradigms. Both paradigms are based on SENet, and only the main encoder and decoder parts of SENet are retained in the illustration. (a) For Paradigm 1, two tasks share the entire network. (b) For Paradigm 2, two tasks share the encoder but have independent decoders.}
      \label{fig:joint_training}
   \end{figure}

\subsection{Loss Function}
\label{sec:loss}

We add image reconstruction as an auxiliary task to the training process, so we use reconstruction loss and segmentation loss to train our SENet together. Next, we introduce these two loss functions respectively.

\vspace{1mm}
\noindent \textbf{Reconstruction loss}. Following ViT \cite{ViT}, we divide the original image $\boldsymbol{I}$ into regular non-overlapping patches. Then we sample a subset from the patches and mask (i.e., remove) the rest (see Sec.~\ref{exp:mask} for the discussion of different masking ratios). Using the encoder operating solely on the sampled patches (without mask tokens) and the decoder to restore the removed ones, we can get the reconstructed image $\boldsymbol{I}_{recon}$. Then we calculate the reconstruction loss between $\boldsymbol{I}_{recon}$ and $\boldsymbol{I}$ with \textit{MSE} loss function:

\begin{equation}
\label{eq2}
\mathcal{L}_{recon}= \textit{MSE}(\boldsymbol{I}_{recon}, \boldsymbol{I}).
\end{equation}

Note that referring to MAE \cite{MAE}, we only calculate the loss of the masked part.

\begin{table}[t]
\centering
\caption{Summary of datasets considered in this work. We show the number of images in training set (Train) and testing set (Test) for different datasets. “-”: the number of images in the training set is $0$, and this dataset is only used for testing.}

\resizebox{0.9\linewidth}{!}{\begin{tabular}{c|c|c|c}
\hline
Task & Dataset   & Train & Test \\ \hline
\multirow{4}{*}{\begin{tabular}[c]{@{}c@{}}Camouflaged\\ Object Detection\end{tabular}} 
& CAMO \cite{CAMO}     & 1\,000  & 250  \\ \cline{2-4} 
& CHAMELEON \cite{CHAMELEON} & -     & 76   \\ \cline{2-4} 
& COD10K \cite{SINet}   & 3\,040  & 2\,026 \\ \cline{2-4} 
& NC4K \cite{NC4K}     & -     & 4\,121 \\ \hline
\multirow{5}{*}{\begin{tabular}[c]{@{}c@{}}Salient\\ Object Detection\end{tabular}}     
& DUTS  \cite{DUTS}    & 10\,553 & 5\,019 \\ \cline{2-4} 
& DUT-OMRON \cite{DUT-OMRON} & -     & 5\,168 \\ \cline{2-4} 
& HKU-IS \cite{HKU-IS}   & -     & 4\,447 \\ \cline{2-4} 
& ECSSD \cite{ECSSD}    & -     & 1\,000 \\ \cline{2-4} 
& PASCAL-S \cite{PASCAL-S} & -     & 850  \\ \hline
\end{tabular}}

\label{table:datasets}
\end{table}

\begin{table*}[t]
\centering
\caption{Quantitative results of our method and other methods on four COD benchmark datasets. “*”: methods proposed for SOD. “$\uparrow / \downarrow$”: the higher/lower the better. “-”: Not available. The top two results are highlighted in \textcolor{red}{red} and \textcolor{blue}{blue}.}
\resizebox{\linewidth}{!}{\begin{tabular}{c|c|cccc|cccc|cccc|cccc}
\hline
\multirow{2}{*}{Method} & \multirow{2}{*}{Venue} 
& \multicolumn{4}{c|}{CAMO-Test \cite{CAMO}}                     
& \multicolumn{4}{c|}{CHAMELEON \cite{CHAMELEON}}                    
& \multicolumn{4}{c|}{COD10K-Test \cite{SINet}}                  
& \multicolumn{4}{c}{NC4K \cite{NC4K}} 
\\ \cline{3-18} 
&                      
&$S_{\alpha}\uparrow$ &$E_{\phi}\uparrow$  &$F_{\beta}^{\omega}\uparrow$  &\multicolumn{1}{c|}{$M \downarrow$}       
&$S_{\alpha}\uparrow$ &$E_{\phi}\uparrow$  &$F_{\beta}^{\omega}\uparrow$  &\multicolumn{1}{c|}{$M \downarrow$}     
&$S_{\alpha}\uparrow$ &$E_{\phi}\uparrow$  &$F_{\beta}^{\omega}\uparrow$  &\multicolumn{1}{c|}{$M \downarrow$}   
&$S_{\alpha}\uparrow$ &$E_{\phi}\uparrow$  &$F_{\beta}^{\omega}\uparrow$    & $M \downarrow$    
\\ \hline

F3Net* \cite{f3net} & AAAI 20             & .711 &.780  &.564  &.109
&.848 &.901 &.744 & .047 &.739 &.819 &.544 &.051 &.780 &.848 &.656 &.070\\

ITSD* \cite{ITSD}  &CVPR 20 &.750 &.830 &.610 
&.102  &.814 &.901 &.662 &.057
&.767 &.861 &.557 &.052 &.811 &.883 &.679 &.064  \\

UCNet* \cite{UCNet} &CVPR 20 &.739  &.820 &.640 &.094
&.880 &.921 &.817 &.036 &.776 &.867 &.633 &.042 &.811 &.886 &.729 &.055 \\

SINet \cite{SINet}  & CVPR 20             & .751 & .771 & .606 & .100 & .869 & .891 & .740 & .044 & .771 & .806 & .551 &.051 & .810 & .873 & .772 & .057 \\

BGNet \cite{BGNet} &IJCAI 21      &  .812     &   .870    &   .749    & .073      &  .901     & .943      &  .850     & .027      & .831      &  .901     &  .722     & .033      &  .851     &    .907   &   .788    & .044      \\

SINet-V2 \cite{SINet-V2} &TPAMI 21   &  .820     & .882      &  .743     & .070     &  .888     &  .942     & .816      & .030       &  .815     & .887      &   .680    & .037     & .847      &  .903     & .769      &  .048     \\

LSR \cite{NC4K}  &CVPR 21    &  .793     &   .826    & .725      & .085      & .893      & .938      & .839      &.033    & .793      &  .868     &   .685    & .041    & .839      &  .883     & .779      &   .053    \\

UJSC \cite{UJSC} &CVPR 21     & .803      &  .853     &  .759     & .076      &  .894     &  .943     &  .848     & .030      &   .817    & .892      & .726      & .035     &  .842     &   .907    &  .771     &  .047     \\

BGSANet \cite{BGSANet} &AAAI 22 
&.796 &.851 &.768 &.079
&.895 &.946 &.851 &.027
&.818 &.891 &.723 &.034
&.841 &.897 &.805 &.048 \\

OSFormer \cite{OSformer} &ECCV 22
&.799 &.858 &.767 &.073
&.891 &.939 &.836 &.028
&.811 &.881 &.701 &.034
&.832 &.891 &.790 &.049 \\

ZoomNet \cite{ZoomNet}   &CVPR 22 &  .820     &  .892     &   .752   & .066     & .902      &  .958     & .845      & .023     &  .838     &   .911    &   .729    & .029      & .853      &  .912     &   .784    &  .043     \\

SegMaR \cite{SegMAR} &CVPR 22          &  .815     &  .872     &  .742     & .071    &  .906     & .954      & .860      &.025     &  .833     &  .895     & .724      & .033      &  .845     &  .892     & .793      &  .047     \\

FPNet \cite{FPNet1}  &MM 23              &  .852     &  \textcolor{blue}{.905}     &  .806     & .056      & \textcolor{blue}{.914}     &  \textcolor{blue}{.961}     &  .856    & \textcolor{blue}{.022}      &  .850     & .913      &  .748   & .029    &  -     &  -     &   -    & -    \\

EVP \cite{EVP} &CVPR 23             & .846      &  .895     &  .777     & .059      &  .871     & .917      &    .795   & .036      &   .843    &  .907     &  .742     & .029    &  -     &   -    &  -     & -      \\

FEDER \cite{FEDER} &CVPR 23         &  .822     & .886      & \textcolor{blue}{.809}     & .067      &   .907    &  \textcolor{red}{.964}     &  \textcolor{blue}{.874}     & .025     &  \textcolor{blue}{.851}     &  \textcolor{blue}{.917}     & \textcolor{blue}{.752}      & .028      &   .863    & \textcolor{blue}{.917}      &   \textcolor{blue}{.827}    &  .042     \\

FSP \cite{FSP}  &CVPR 23            &  \textcolor{blue}{.856}     &  .899     &  .799     & \textcolor{blue}{.050}      & .908      &  .942     &  .851     &.023     &  \textcolor{blue}{.851}     & .895      &  .735     & \textcolor{blue}{.026}      &  \textcolor{blue}{.879}     &  .915     &   .816    & \textcolor{blue}{.035}      \\ 
\hline

\textbf{SENet (Ours)}      &-                    
& \textcolor{red}{.888}  &  \textcolor{red}{.932}   &\textcolor{red}{.847}  &\textcolor{red}{.039}
& \textcolor{red}{.918}  &  .957    &  \textcolor{red}{.878}   & \textcolor{red}{.019} 
& \textcolor{red}{.865}  &  \textcolor{red}{.925}    &\textcolor{red}{.780} &\textcolor{red}{.024} 
&\textcolor{red}{.889}  & \textcolor{red}{.933}  & \textcolor{red}{.843}  & \textcolor{red}{.032} 
\\ \hline

\end{tabular}}
\label{table:codsota}

\end{table*}

\vspace{1mm}
\noindent \textbf{Segmentation loss}. Previous works \cite{ZoomNet, MSIN} have pointed out that the smaller the size of the target object, the more difficult it is to complete the COD and SOD tasks. Therefore, to make the network pay more attention to those difficult segmentation instances with smaller target objects, we propose the dynamic weighted loss (DW loss) based on BCE and IoU loss. 

Specifically, we set different weights $\alpha$ for the edge areas of the image according to the size of the target objects, 
\begin{equation}
\label{eq3}
\alpha = l \cdot \frac{S_{img}}{S_{obj}},
\end{equation}

where $S_{img}$ is the area of the image (Height $\times$ Width), and $S_{obj}$ is the area of the target region, which can be easily obtained by counting the number of $1$ in the ground truth binary image. $l$ is a hyperparameter, which is set to $1$. 

The weights of the remaining regions are all set to $1$. We only set the weight of the boundary region between the foreground and background to $\alpha$ to highlight the importance of the edge area, hoping to better use edge information to guide the network to complete segmentation without extra edge annotation information as input to the network \cite{BGNet}. In this way, we get the weight matrix $\boldsymbol{W}$, $\boldsymbol{W}$ is consistent with the size of the input image, and different weights are set for pixels in different areas when calculating segmentation loss. 

In addition, due to the uncertainty in the annotation process \cite{UJSC-V2}, \footnote{Some images are of poor quality or the objects are highly camouflaged, causing the human eye to be unable to distinguish the boundaries between the target region and the non-target region, resulting in labeling uncertainty at the edge areas.} we use bilinear interpolation method to set new ground truth for pixels in the edge area. In this way, we obtain ground truth image $\boldsymbol{I}_{gt}$, containing more than $0$ and $1$ ($1$ for foreground, $0$ for background, and interpolation between $0$ and $1$ for the transition area between the two), this can also distinguish the three areas, without the need of additional annotation to help us find the edge area \cite{egnet}. 

As in most previous works \cite{BGNet,FEDER}, we use BCE and IoU loss to calculate the segmentation loss between the prediction map $\boldsymbol{I}_{pred}$ generated by SENet and the ground truth $\boldsymbol{I}_{gt}$, using $\boldsymbol{W}$ for weighting:

\begin{equation}
\label{eq4}
\mathcal{L}_{seg}= \boldsymbol{W} \odot (\textit{BCE}(\boldsymbol{I}_{pred}, \boldsymbol{I}_{gt}) + \textit{IoU}(\boldsymbol{I}_{pred}, \boldsymbol{I}_{gt})),
\end{equation}
where $\odot$ indicates pixel-wise multiplication and then average. 


\vspace{1mm}
\noindent \textbf{Total loss}. We train our SENet with the above reconstruction loss and segmentation loss together:

\begin{equation}
\label{eq5}
\mathcal{L}_{total}= \lambda \cdot \mathcal{L}_{recon} + (1-\lambda) \cdot \mathcal{L}_{seg},
\end{equation}
where $\lambda$ is a hyperparameter, which is set to $0.1$ in this work (lower weight for auxiliary image reconstruction task) .

\subsection{Joint Training}
\label{sec:joint_training}

Previous works \cite{UJSC, UJSC-V2} have pointed out that there is a strong contradiction between SOD and COD tasks, because the former focuses on salient objects in the image while the latter focuses on disguised inconspicuous objects. One natural question is how the proposed SENet performs in joint training (i.e., using COD and SOD data to train the network together) for SENet is a general architecture for both tasks.

To study it, we propose two paradigms for joint training, as shown in Fig.~\ref{fig:joint_training}. Paradigm 1 simply mixes COD and SOD data for training, which is equivalent to treating the two tasks as the same binary segmentation task, segmenting the input image into the target region and the non-target region. Paradigm 2 uses a shared encoder structure, using two independent decoders for two tasks. This design idea comes from the common practice of sharing upstream backbone between different downstream tasks in multi-task learning \cite{MTL1, MTL2}. We believe that the features extracted by the encoder are universal, while the two decoders are used to Algorithm 1 clearly shows the process of joint trainingsegment salient objects and camouflaged objects respectively. The shared encoder seeks common knowledge and two independent decoders store difference. 

We have actually tried some other structures (e.g., sharing all the front layers and only using different $1 \times 1$ convolution layers), but the performance is poor so we do not introduce them in detail.

 Algorithm~\ref{algorithm:joint_training} clearly shows the process of joint training. It can be seen that we train the network using both COD and SOD data simultaneously. We calculate the loss separately for each of the two forward processes, and then, based on the average of these two loss, we perform backward propagation to compute gradients and optimize the parameters for each respective component.

\section{Experiments}
\label{sec:experiment}

\begin{table*}[t]
\centering
\caption{Quantitative results of our method and other methods on five SOD benchmark datasets. “*”: methods proposed for COD. The top two results are highlighted in \textcolor{red}{red} and \textcolor{blue}{blue}.}
\resizebox{\linewidth}{!}{\begin{tabular}{c|c|cccc|cccc|cccc|cccc|cccc}
\hline
\multirow{2}{*}{Method} & \multirow{2}{*}{Venue} 
& \multicolumn{4}{c|}{DUTS-Test \cite{DUTS}}                          
& \multicolumn{4}{c|}{DUT-OMRON \cite{DUT-OMRON}}                     
& \multicolumn{4}{c|}{HKU-IS \cite{HKU-IS}}                        
& \multicolumn{4}{c|}{ECSSD \cite{ECSSD}}                          
& \multicolumn{4}{c}{PASCAL-S \cite{PASCAL-S}}  \\ \cline{3-22} 
                        &        
&$S_{\alpha}\uparrow$ &$E_{\phi}\uparrow$  &$F_{\beta}^{m}\uparrow$  &$M \downarrow$ 
&$S_{\alpha}\uparrow$ &$E_{\phi}\uparrow$  &$F_{\beta}^{m}\uparrow$  &$M \downarrow$ 
&$S_{\alpha}\uparrow$ &$E_{\phi}\uparrow$  &$F_{\beta}^{m}\uparrow$  &$M \downarrow$ 
&$S_{\alpha}\uparrow$ &$E_{\phi}\uparrow$  &$F_{\beta}^{m}\uparrow$  &$M \downarrow$ 
&$S_{\alpha}\uparrow$ &$E_{\phi}\uparrow$  &$F_{\beta}^{m}\uparrow$  &$M \downarrow$ \\ 
\hline

F3Net \cite{f3net}  &AAAI 20
&.888 &.902 &.840 &.035
&.838 &.870 &.766 &.053
&.917 &.953 &.910 &.028
&.924 &.927 &.925 &.033
&.855 &.859 &.840 &.062  \\

MINet \cite{MINet}  &CVPR 20
&.884 &.917 &.884 &.037
&.833 &.873 &.810 &.055
&.920 &.961 &.935 &.028
&.925 &.953 &.947 &.033
&.857 &.899 &.882 &.064  \\

LDF \cite{LDF}                      &   CVPR 20             
&  .892     &  .925     & .861      &   .034 
&   .839     & .865      & .770      &  .052  
&   .920    &  .953     &   .913    &  .028  
& .919      & .943      & 923      &  .036     
&  .860     & .901      &  .856     & .063      \\

SINet* \cite{SINet}       &CVPR 20
&.872    &.904    &.846    &.042
&.825    &.851    &.757    &.058
&.911    &.942    &.916    &.033
&.916    &.939    &.929    &.041
& .855   &.888    &.842    &.069     \\                         

BGNet* \cite{BGNet} &IJCAI 21
&.891 &.922 &.873 &.033
&.834 &.858 &.769 &.050
&.917 &.951 &.926 &.028
&.930 &.954 &.942 &.029
&.862 &.900 &.849 &.058 \\

VST \cite{VST}                      & ICCV 21         
& .896      & .892      & .890      &  .037    
&  .850      & .861      & .825      & .058     
&  .928     &.953       & .942      &  .029    
&  .932     & .918      &  .951     & .033     
& .865      & .837      &  .875     & .061      \\

UJSC \cite{UJSC}    & CVPR 21           
& .899      &  .937     & .866      & .032      
& .850       & .884      &  .782     &  .051    
& .931     &  \textcolor{blue}{.967}     & .924      & .026     
& .933     &  .960     & .935      & .030     
& .864    & .902      &  .841     &  .062     \\

SelfReformer \cite{SelfReformer}              &   TMM 22                   
& .911      & .920      &  .916     &  .026    
&  .856      &  .886     & .836      &  \textcolor{blue}{.041}    
&  .930     & .959      & .947      & .024   
&   .935    &  .928     &  .957     &  .027    
&  .874     & .872      &  \textcolor{red}{.894}     &  .050     \\

UPL \cite{UPL}                       &  AAAI 22                     
& .846      & -      &.783       & .050     
&  .808      &  -     & .711      & .059     
& .897      &-       & .879      & .035     
& .899      & -      & .885      & .043     
& .822      &  -     &  .773     &   .080    \\

PSOD \cite{PSOD}                       &  AAAI 22                
&  .853     & -      & .858      &  .045  
&  .824      & -      & .809      &   .064 
&  .902     &  -     &   .923    &   .032  
&  .914     & -      &   .936    &   .035 
& .853      & -      &   .866    &  .064     \\

ZoomNet* \cite{ZoomNet}    &CVPR 22
&.900 &.936 &.866 &.033
&.841 &.872 &.771 &.053
&.931 &\textcolor{blue}{.967} &.923 &.023
&.935 &.963 &.933 &.027
&.869 &\textcolor{blue}{.917} &.860 &.057   \\

BBRF \cite{BBRF}                 &     TIP 23                 
& .908      &  .927     & .916      &   \textcolor{blue}{.025 }   
& .855       & .887      & .843      &  .042    
& .935      &  .965     &  \textcolor{red}{.958}     & \textcolor{blue}{ .020 }   
&  .939     & .934      &  .963     & \textcolor{blue}{ .022}     
&  .871     &  .867     &  \textcolor{blue}{.891}     & \textcolor{blue}{ .049}    \\

TGSD \cite{TGSD}                      &CVPR 23               
&.842       & .902      & .824      &   .047
& .812       &.863       & .763      &   .061 
&  .890     & .942      &  .906     &    .036 
&  .894     &  .937     &  .918     &  .044  
&   .830    & .887      & .828      &   .072    \\

EVP \cite{EVP}     &   CVPR 23                    
& .913 & \textcolor{blue}{.947} &\textcolor{blue}{.923} & .026 
& .862  & .894 & \textcolor{blue}{.858} & .046 
& .931 & .961 & .952 & .024 
& .935 & .957 & .960 & .027 
& \textcolor{blue}{.878} & \textcolor{blue}{.917} & .872 & .054 \\

UJSC-V2 \cite{UJSC-V2}             & arXiv 23          
&  .900     & .937      &  .875     & .030
&  .841     & .876      & .777       &.050 
& .921     &.958       & .920     & .026  
&   .929    & .955      &.935      &  .029  
&   .866    &.910     & .867      &  .058     \\

EVP-V2 \cite{EVP-V2}             & arXiv 23          
&  \textcolor{blue}{.915}     & .944      &  .921     & .026
& \textcolor{blue}{.874}     & \textcolor{red}{.902}      & \textcolor{red}{.865}  &\textcolor{red}{.040} 
&  \textcolor{blue}{.937}     &.966       & \textcolor{blue}{.957}     & \textcolor{blue}{.020}  
&   \textcolor{blue}{.944}    & \textcolor{blue}{.964 }     & \textcolor{red}{.967}      & \textcolor{blue}{ .022  }
&   .877    &.912     & .868      &  .054     \\

\hline
\textbf{SENet (Ours)}                    &  -
&\textcolor{red}{.926} &\textcolor{red}{.953} &\textcolor{red}{.925}  &\textcolor{red}{.022}
&\textcolor{red}{.876} &\textcolor{blue}{.899}      & .835      & \textcolor{red}{.040}
&\textcolor{red}{.941} &\textcolor{red}{.970}  & .953 &\textcolor{red}{.019} 
&\textcolor{red}{.949} &\textcolor{red}{.968}   & \textcolor{blue}{.964}     &  \textcolor{red}{.020}
&\textcolor{red}{.890} &\textcolor{red}{.923}  &.886   & \textcolor{red}{.046}     \\ \hline

\end{tabular}}

\label{table:sodsota}
\end{table*}

\begin{figure*}[t]
      \centering
     
      \includegraphics[width=1.0\linewidth]{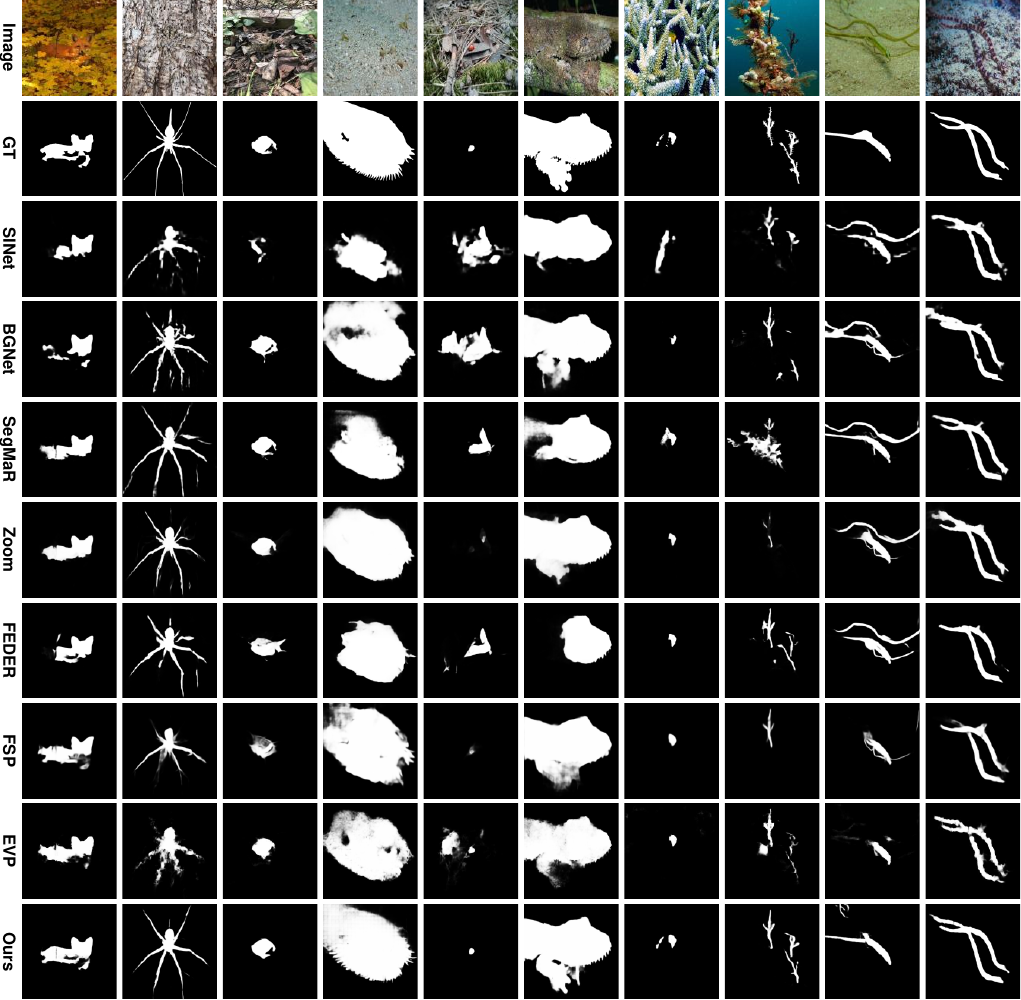}
       \caption{\small Visual comparison of our camouflage predictions with the state-of-the-art (SOTA) methods on different types of COD samples. Please zoom in for more details.}
      \label{fig:cod_results}
   \end{figure*}

\begin{figure*}[t]
      \centering
       
      \includegraphics[width=1.0\linewidth]{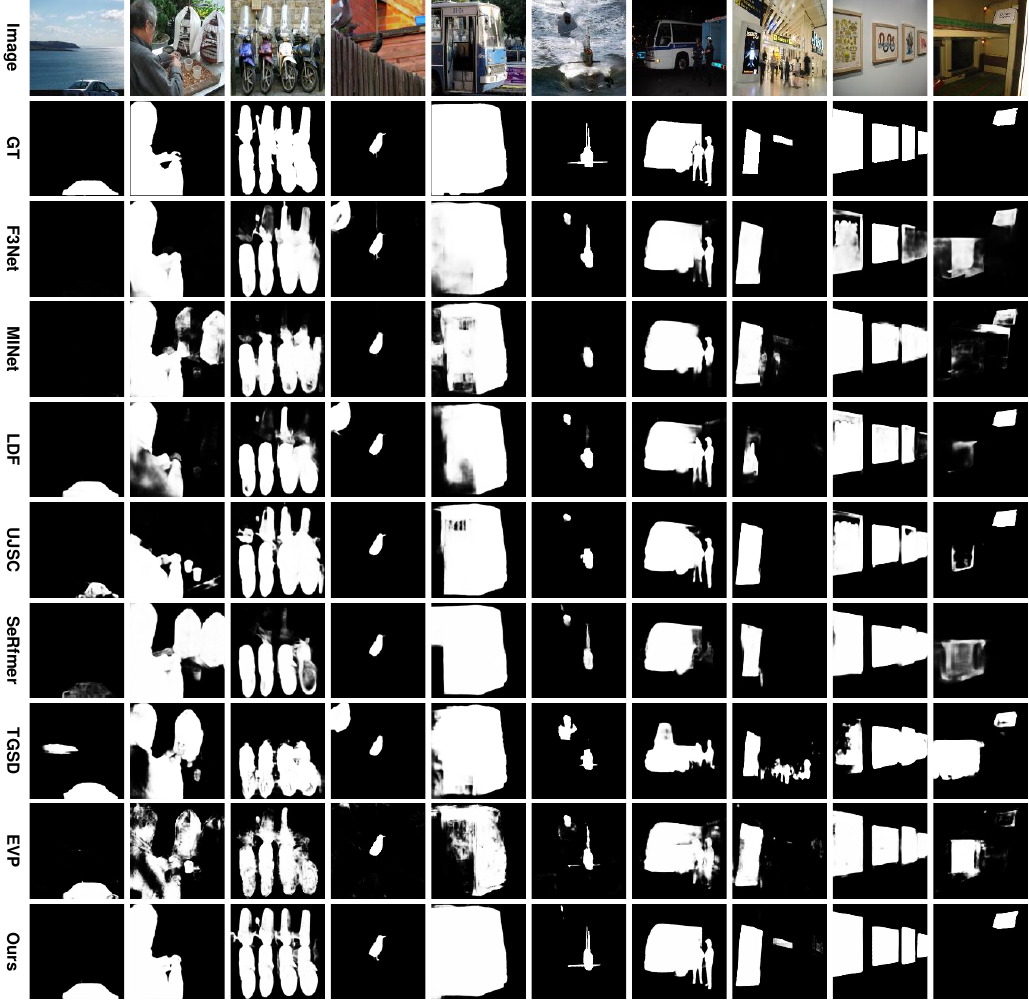}
     \caption{\small Visual comparison of our saliency predictions with the SOTA methods on different types of SOD samples. Please zoom in for more details.}
      \label{fig:sod_results}
   \end{figure*}

\begin{table*}[t]
\centering
\caption{Comparison of the computational efficiency of different masking ratios. The time used here is the time it takes to complete the COD task training on $1$ V100 GPU for $1$ epoch.}
\resizebox{\linewidth}{!}{
\begin{tabular}{c|cccccccccccccccccccc}
\hline
Masking ratio ($\%$)     
& 0      & 5      & 10     & 15     & 20   
& 25     & 30    & 35     & 40     & 45    
& 50    & 55     & 60     & 65     & 70    
& 75     & 80     & 85    & 90     & 95     \\ \hline

GFLOPS                 
& 88.5   & 85.2   & 81.9   & 78.6   & 75.3  
& 72.1   & 68.8  & 65.4   & 62.1   & 58.8   
& 55.6  & 52.3   & 49.0   & 45.7   & 42.4   
& 39.2   & 35.8   & 32.5  & 29.2   & 25.9   \\ \hline

Time 
& $4^\prime 46^{\prime \prime}$ 
& $4^\prime 35^{\prime \prime}$ 
& $4^\prime 26^{\prime \prime}$ 
& $4^\prime 18^{\prime \prime}$ 
& $4^\prime 8^{\prime \prime}$ 
& $3^\prime 58^{\prime \prime}$ 
& $3^\prime 44^{\prime \prime}$ 
& $3^\prime 35^{\prime \prime}$ 
& $3^\prime 27^{\prime \prime}$ 
& $3^\prime 19^{\prime \prime}$ 
& $3^\prime 6^{\prime \prime}$ 
& $2^\prime 57^{\prime \prime}$ 
& $2^\prime 48^{\prime \prime}$ 
& $2^\prime 40^{\prime \prime}$ 
& $2^\prime 33^{\prime \prime}$ 
& $2^\prime 28^{\prime \prime}$
& $2^\prime 21^{\prime \prime}$ 
& $2^\prime 5^{\prime \prime}$ 
& $1^\prime 58^{\prime \prime}$ 
& $1^\prime 49^{\prime \prime}$ \\ \hline

\end{tabular}}

\label{table:efficiency}
\end{table*}
\begin{table}[t]
\centering
\caption{Performance comparison of different training methods. “COD Train”: the model trained on COD training set. “SOD Train”: the model trained on SOD training set. “Paragigm 1/ 2”: model trained through joint training Paradigm 1/ 2. The top two results are highlighted in \textcolor{red}{red} and \textcolor{blue}{blue}. \underline{Underline} denotes the results of cross-domain evaluation (training on COD, testing on SOD or training on SOD, testing on COD), it can be seen clearly that this is also the worst performance among the four cases.}
\resizebox{\linewidth}{!}{
\begin{tabular}{cc|cccc}
\hline
\multicolumn{2}{c|}{Dataset \& Metric}               & COD Train & SOD Train & Paradigm 1    & Paradigm 2 \\ \hline
\multicolumn{1}{c}{\multirow{4}{*}{CAMO-Test}}      &$S_{\alpha}\uparrow$ 
&\textcolor{red}{.888}  &\underline{.734}       &.853       &\textcolor{blue}{.861}      \\
\multicolumn{1}{c}{}                           &$E_{\phi}\uparrow$   
&\textcolor{red}{.932}  &\underline{.742}       &.907       &\textcolor{blue}{.915}      \\
\multicolumn{1}{c}{}                           &$F_{\beta}^{\omega}\uparrow$ &\textcolor{red}{.847}  &\underline{.608}       &.808       &\textcolor{blue}{.822}      \\
\multicolumn{1}{c}{}                           & $M \downarrow$  
&\textcolor{red}{.039}  &\underline{.101}       &.052       &\textcolor{blue}{.046}     \\ \hline
\multicolumn{1}{c}{\multirow{4}{*}{CHEMELEON}} &$S_{\alpha}\uparrow$ 
&\textcolor{red}{.918}  &\underline{.743}       &.890       &\textcolor{blue}{.896}      \\
\multicolumn{1}{c}{}                           &$E_{\phi}\uparrow$  
&\textcolor{red}{.957}  &\underline{.747}       &.939       &\textcolor{blue}{.943}      \\
\multicolumn{1}{c}{}                           &$F_{\beta}^{\omega}\uparrow$ &\textcolor{red}{.878}  &\underline{.590}       &.840       &\textcolor{blue}{.851}      \\
\multicolumn{1}{c}{}                           & $M \downarrow$  
&\textcolor{red}{.019}  &\underline{.074}       &.030       &\textcolor{blue}{.023}      \\ \hline
\multicolumn{1}{c}{\multirow{4}{*}{COD10K-Test}}    &$S_{\alpha}\uparrow$ 
&\textcolor{red}{.865}  &\underline{.751}       &.841       &\textcolor{blue}{.849}      \\
\multicolumn{1}{c}{}                           &$E_{\phi}\uparrow$   
&\textcolor{red}{.925}  &\underline{.769}       &.901       &\textcolor{blue}{.916}      \\
\multicolumn{1}{c}{}                           &$F_{\beta}^{\omega}\uparrow$ &\textcolor{red}{.780}  &\underline{.583}       &.739       &\textcolor{blue}{.760}      \\
\multicolumn{1}{c}{}                           & $M \downarrow$  
&\textcolor{red}{.024}  &\underline{.059}       &.032       &\textcolor{blue}{.027}      \\ \hline
\multicolumn{1}{c}{\multirow{4}{*}{NC4K}}      &$S_{\alpha}\uparrow$ 
&\textcolor{red}{.889}  &\underline{.814}       &.874       &\textcolor{blue}{.878}      \\
\multicolumn{1}{c}{}                           &$E_{\phi}\uparrow$   
&\textcolor{red}{.933}  &\underline{.835}       &.919       &\textcolor{blue}{.928}      \\
\multicolumn{1}{c}{}                           &$F_{\beta}^{\omega}\uparrow$ &\textcolor{red}{.843}  &\underline{.722}       &.825       &\textcolor{blue}{.836}      \\
\multicolumn{1}{c}{}                           & $M \downarrow$ 
&\textcolor{red}{.032}  &\underline{.061}       &.039       &\textcolor{blue}{.034}      \\ \hline
\multicolumn{1}{c}{\multirow{4}{*}{DUTS-Test}}   &$S_{\alpha}\uparrow$ 
&\underline{.867}  &.926       &\textcolor{red}{.932}       &\textcolor{blue}{.929}      \\
\multicolumn{1}{c}{}                           &$E_{\phi}\uparrow$   
&\underline{.892}  &.953       &\textcolor{red}{.959}       &\textcolor{blue}{.957}      \\
\multicolumn{1}{c}{}                           &$F_{\beta}^{m}\uparrow$ 
&\underline{.846}  &.925       &\textcolor{red}{.931}       &\textcolor{blue}{.930}      \\
\multicolumn{1}{c}{}                          & $M \downarrow$ 
&\underline{.046}  &.022       &\textcolor{red}{.020}       &\textcolor{blue}{.021}      \\ \hline
\multicolumn{1}{c}{\multirow{4}{*}{DUT-OMRON}} &$S_{\alpha}\uparrow$ 
&\underline{.699}  &\textcolor{blue}{.876}       &\textcolor{red}{.878}       &.874      \\
\multicolumn{1}{c}{}                           &$E_{\phi}\uparrow$   
&\underline{.710}  &.899       &\textcolor{red}{.907}       &\textcolor{blue}{.901}      \\
\multicolumn{1}{c}{}                          &$F_{\beta}^{m}\uparrow$ 
&\underline{.637}  &.835       &\textcolor{blue}{.839}       &\textcolor{red}{.844}      \\
\multicolumn{1}{c}{}                           & $M \downarrow$  
&\underline{.096}  &\textcolor{blue}{.040}       &\textcolor{red}{.039}       &.048      \\ \hline
\multicolumn{1}{c}{\multirow{4}{*}{HKU-IS}}    &$S_{\alpha}\uparrow$ 
&\underline{.808}  &\textcolor{red}{.941}       &.939       &\textcolor{blue}{.940}      \\
\multicolumn{1}{c}{}                           &$E_{\phi}\uparrow$   
&\underline{.823}  &\textcolor{red}{.970}       &.968       &\textcolor{red}{.970}      \\
\multicolumn{1}{c}{}                           &$F_{\beta}^{m}\uparrow$ 
&\underline{.809}  &\textcolor{red}{.953}       &.949       &\textcolor{blue}{.952}      \\
\multicolumn{1}{c}{}                           & $M \downarrow$  
&\underline{.072}  &\textcolor{red}{.019}       &.020       &\textcolor{red}{.019}      \\ \hline
\multicolumn{1}{c}{\multirow{4}{*}{ECSSD}}     &$S_{\alpha}\uparrow$ 
&\underline{.805}  &\textcolor{blue}{.949}       &.946       &\textcolor{red}{.950}      \\
\multicolumn{1}{c}{}                           &$E_{\phi}\uparrow$   
&\underline{.812}  &\textcolor{blue}{.968}       &.966       &\textcolor{red}{.970}      \\
\multicolumn{1}{c}{}                           &$F_{\beta}^{m}\uparrow$ 
&\underline{.821}  &\textcolor{blue}{.964}       &.958       &\textcolor{red}{.966}      \\
\multicolumn{1}{c}{}                          & $M \downarrow$  
&\underline{.087}  &\textcolor{blue}{.020}       &.022       &\textcolor{red}{.019}      \\ \hline
\multicolumn{1}{c}{\multirow{4}{*}{PASCAL-S}}  &$S_{\alpha}\uparrow$ 
&\underline{.799}  &\textcolor{blue}{.890}       &.887       &\textcolor{red}{.891}      \\
\multicolumn{1}{c}{}                           &$E_{\phi}\uparrow$   
&\underline{.823}  &\textcolor{blue}{.923}       &.921       &\textcolor{red}{.927}      \\
\multicolumn{1}{c}{}                           &$F_{\beta}^{m}\uparrow$ 
&\underline{.803}  &\textcolor{blue}{.886}       &.882       &\textcolor{red}{.892}      \\
\multicolumn{1}{c}{}                           & $M \downarrow$ 
&\underline{.091}  &\textcolor{red}{.046}       &.047       &\textcolor{red}{.046}      \\ \hline
\end{tabular}
}

\label{table:joint}
\end{table}
\begin{table*}[t]
\centering
\caption{Ablation study on two COD datasets and two SOD datasets. “w/o DW loss” indicates not using the proposed DW loss, using ordinary BCE and IoU loss instead, which is unweighted. “Nearest Interpolation” means to use nearest interpolation method when resizing the ground truth binary image to the specified size, bilinear interpolation is used by default in this work. “Pixel Position Aware loss \cite{f3net}” is another widely used weighted BCE and IoU loss. “ViT Pretrained Encoder” and “Randomly Initialized Decoder” represent using two different initializations to initialize SENet, the MAE \cite{MAE} pre-trained encoder and decoder are used by default. The above changes or deletions are all for the default SENet, and other settings are consistent with the default. Default SENet settings are marked in \colorbox{gray}{gray}.}
\resizebox{\linewidth}{!}{\begin{tabular}{cccccccccccccccccc}
\hline
\multicolumn{1}{c|}{\multirow{2}{*}{Settings}} 
& \multicolumn{4}{c|}{COD10K-Test}                     
& \multicolumn{4}{c|}{NC4K}                    
& \multicolumn{4}{c|}{DUTS-Test}                  
& \multicolumn{4}{c}{HKU-IS}      \\ \cline{2-17} 
\multicolumn{1}{c|}{}              
&$S_{\alpha}\uparrow$ &$E_{\phi}\uparrow$  &$F_{\beta}^{\omega}\uparrow$  &\multicolumn{1}{c|}{$M \downarrow$}       
&$S_{\alpha}\uparrow$ &$E_{\phi}\uparrow$  &$F_{\beta}^{\omega}\uparrow$  &\multicolumn{1}{c|}{$M \downarrow$}     
&$S_{\alpha}\uparrow$ &$E_{\phi}\uparrow$  &$F_{\beta}^{m}\uparrow$  &\multicolumn{1}{c|}{$M \downarrow$}   
&$S_{\alpha}\uparrow$ &$E_{\phi}\uparrow$  &$F_{\beta}^{m}\uparrow$    & $M \downarrow$     \\ \hline
\multicolumn{1}{l|}{w/o LICM \& w/o DW loss}               
&.838  &.901  &.751  & \multicolumn{1}{c|}{.029} 
&.858  &.907  &.811  & \multicolumn{1}{c|}{.038} 
&.905  &.932  &.910  & \multicolumn{1}{c|}{.026} 
&.926  &.944  &.940  &.022  \\
 
\multicolumn{1}{l|}{w/o LICM}              
&.850  &.908  &.756  & \multicolumn{1}{c|}{.028} 
&.874  &.916  &.827  & \multicolumn{1}{c|}{.035} 
&.915  &.943  &.911  & \multicolumn{1}{c|}{.026} 
&.930  &.961  &.940  &.021  \\

\multicolumn{1}{l|}{w/o DW loss}               
&.849  &.911  &.764  & \multicolumn{1}{c|}{.026} 
&.871  &.911  &.824  & \multicolumn{1}{c|}{.036} 
&.914  &.945  &.912  & \multicolumn{1}{c|}{.025} 
&.930  &.959  &.944  &.021  \\
\hline

\multicolumn{1}{l|}{Nearest Interpolation}               
&.857  &.919  &.770  & \multicolumn{1}{c|}{.027} 
&.881  &.926  &.835  & \multicolumn{1}{c|}{.035} 
&.925  &.953  &.920  & \multicolumn{1}{c|}{.024} 
&.939  &.968  &.950  &.020  \\

\multicolumn{1}{l|}{Pixel Position Aware loss \cite{f3net}}               
&.862  &.920  &.778  & \multicolumn{1}{c|}{.026} 
&.889  &.925  &.840  & \multicolumn{1}{c|}{.034} 
&.918  &.954  &.923  & \multicolumn{1}{c|}{.022} 
&.938  &.961  &.948  &.021  \\

 \hline

 \multicolumn{1}{l|}{ViT \cite{ViT} Pretrained Encoder }               
 &.817  &.906  &.698  & \multicolumn{1}{c|}{.033}
 &.856  &.918  &.794  & \multicolumn{1}{c|}{.041}
 &.910  &.946  &.904  & \multicolumn{1}{c|}{.027} 
 &.931  &.965  &.943  &.023  \\

\multicolumn{1}{l|}{Randomly Initialized Decoder}
&.824  &.853  &.671  & \multicolumn{1}{c|}{.037}
&.855  &.877  &.765  & \multicolumn{1}{c|}{.049} 
&.921  &.952  &.917  & \multicolumn{1}{c|}{.024} 
&.937 &.966  &.948  &.022  \\
 \hline

\rowcolor{Gray} \multicolumn{1}{l|}{\textbf{SENet}}
&.865  &.925  &.780  & \multicolumn{1}{c|}{.024} 
&.889  &.933  &.843  & \multicolumn{1}{c|}{.032} 
&.926  &.953  &.925  & \multicolumn{1}{c|}{.022} 
&.941  &.970  &.953  &.019  \\
 \hline
 
\end{tabular}}

\label{tab:ablation}

\end{table*}

\begin{algorithm}[t]
		\caption{Paradigm 2 of joint training}
		{ \small{1}\,\,\,:} {\bfseries for } cod\_image, cod\_gt, sod\_image, sod\_gt in loader:  \\
  { \small{2}\,\,\,:} \ \ \ Calculate the loss of COD part:  \\
		{ \small{3}\,\,\,:} \ \ \ \ \ \,cod\_pred\,=\,model(cod\_image, decoder=‘cod') \\
  { \small{4}\,\,\,:} \ \ \ \ \ \,cod\_loss\,=\,loss\_fn(cod\_pred, cod\_gt) \\
  { \small{5}\,\,\,:} \ \ \ Calculate the loss of SOD part:  \\
  { \small{6}\,\,\,:} \ \ \ \ \ \,sod\_pred\,=\,model(sod\_image, decoder=‘sod') \\
  { \small{7}\,\,\,:} \ \ \ \ \ \,sod\_loss\,=\,loss\_fn(sod\_pred, sod\_gt) \\
    { \small{8}\,\,\,:} \ \ \ Calculate total loss:  \\
    { \small{9}\,\,\,:} \ \ \ \ \ \, loss\,=\,(cod\_loss\,+\,sod\_loss)\,/\,2 \\
     { \small{10}\,\,\,:} \ \ Calculate gradients and update parameters:  \\
     { \small{11}\,\,\,:} \ \ \ \ \, loss.backward() \\
     { \small{12}\,\,\,:} \ \ \  \ \, optimizer.step() \\
		{ \small{13}\,\,\,:} {\bfseries end}
	
		\label{algorithm:joint_training}
	
	\end{algorithm}

\subsection{Experimental Setup}
\label{exp:setup}

\noindent \textbf{Datasets}. We conduct experiments on nine most widely used datasets for COD \cite{CAMO, CHAMELEON, SINet, NC4K} and SOD \cite{DUTS, DUT-OMRON, HKU-IS, ECSSD, PASCAL-S} tasks, the specific information of the datasets is shown in \ref{table:datasets}. Following  \cite{EVP-V2}, for COD, we form the training set with $3\,040$ images from COD10K-Train and $1\,000$ images from CAMO-Train, while the remaining camouflaged images are used for testing, for SOD, we form the training set with $10\,553$ images from DUTS-Train , while the remaining salient images are used for testing. For joint training, we use the training sets of COD and SOD to train the model together and test the performance of this model on all testing sets.

\vspace{1mm}
\noindent \textbf{Metrics}. Following \cite{EVP-V2}, structure measure  ($S_{\alpha}$), mean E-measure ($E_{\phi}$), F-measure ($F_{\beta}$), and mean absolute error ($M$) are reported to evaluate the segmentation performance. Note that weighted F-measure ($F_{\beta}^{\omega}$) is used for COD and maximum F-measure ($F_{\beta}^{m}$) is used for SOD. In addition, to more quickly compare the performance of different methods instead of comparing the above four indicators one by one, we propose a segmentation score (the higher the better):

\begin{equation}
\label{eq6}
\textit{Score} = S_{\alpha} + E_{\phi} + F_{\beta} + (1-M) .
\end{equation}

We propose this rough metric based on the fact that in most cases one method is better than another and will be better on all four metrics, and we only use this metric in Fig.~\ref{fig:radar} and Fig.~\ref{fig:mask_ratio}, for the convenience of drawing. 

\vspace{1mm}
\noindent \textbf{Implementation details}. The proposed SENet is implemented in PyTorch on one V100 GPU and is optimized by the Adam. For the settings of the ``dim'', ``head'' and “depth” parameters in the Transformer block, for encoder, ``dim'' is set to $768$, ``head'' is $12$ and “depth” is $12$, for decoder, ``dim'' is set to $512$, ``head'' is $16$ and “depth” is $8$, and use the MAE \cite{MAE} pre-trained encoder and decoder weights to initialize. We resize all the input images to $384 \times 384$ and augment them by randomly horizontal flipping. The masking ratio is set to $5\%$ during training and $0$ during testing. All parameters are updated during the training process of $25$ epochs, the learning rate is initialized to $10^{-4}$ and adjusted by poly strategy with the power of $0.9$.

\subsection{Performance Comparison on COD}
\label{exp:codsota}

\noindent \textbf{Quantitative analysis}. Table~\ref{table:codsota} summarizes the quantitative results of our SENet against $16$ competitors on four challenging COD benchmark datasets under four evaluation metrics. It can be seen clearly that those methods \cite{f3net, ITSD, UCNet} designed for SOD all perform poorly on COD task. Furthermore, our method achieves the best results in $15$ out of $16$ comparisons, especially in the comparison of the most difficult CAMO-Test dataset, which is significantly ahead of the other methods (e.g., the $M$ indicator is $0.011$ smaller than the second place). The indicators on these datasets are already very high, and more difficult datasets can reflect the strength of our method. The remaining three datasets are relatively less difficult \cite{UJSC-V2}, but our SENet has also achieved different degrees of leadership.

\vspace{1mm}
\noindent \textbf{Qualitative analysis}.  Fig.~\ref{fig:cod_results} displays a visual comparison of SENet with some representative competitors in several typical camouflage scenarios, including small, multiple, complex-shaped, and blurred boundary disguises \cite{FSP}. It can be seen that the compared methods \cite{SINet, BGNet, SegMAR, ZoomNet, FEDER, FSP, EVP} tend to provide incorrect shapes, incomplete target regions, or even miss the target altogether, resulting in inferior segmentation performance for camouflaged objects. Our proposed method exhibits superior visual performance by delivering more accurate and complete predictions.

\subsection{Performance Comparison on SOD}
\label{exp:sodsota}

\noindent \textbf{Quantitative analysis}. Table~\ref{table:sodsota} summarizes the quantitative results of our SENet against $16$ competitors on five challenging SOD benchmark datasets under four evaluation metrics. It can be seen that those methods \cite{SINet, ZoomNet, BGNet} designed for COD also have not bad performance on SOD task, but it is much inferior compared to their strong performance on COD task. Furthermore, our method achieves the best results in $15$ out of $20$ comparisons and further improves some already high indicators. Since SOD task is relatively simpler \cite{SelfReformer}, we are not as far ahead as we are on COD task, but this has also proven that our simple architecture model can achieve the same or even better results than carefully designed ones.

\begin{figure}[t]
      \centering
      \includegraphics[width=1.0\linewidth]{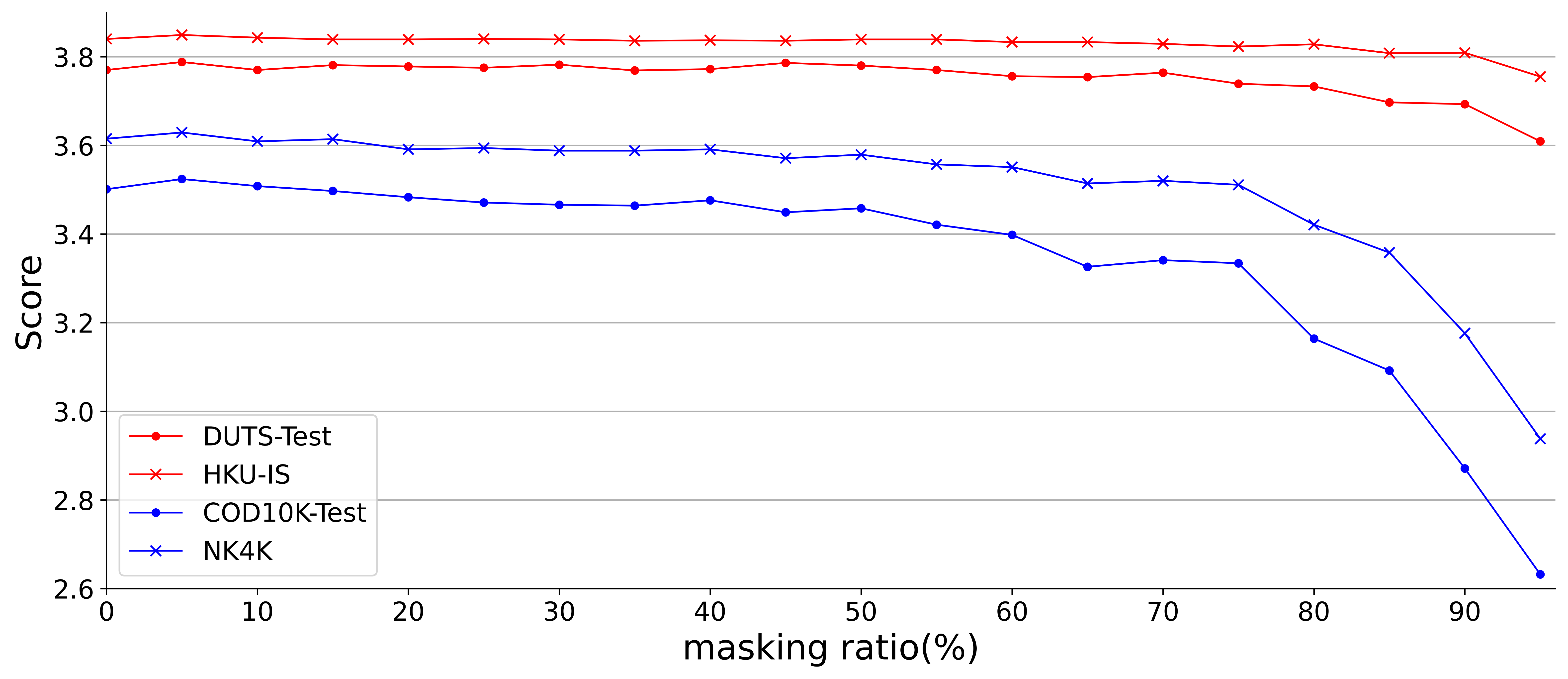}
      \caption{\small The impact of training SENet with different masking ratios on its performance on two COD datasets and two SOD datasets. The two \textcolor{red}{red lines} above are for SOD datasets, and the two \textcolor{blue}{blue lines} below are for COD datasets. For the evaluation metric used here, see Eq.~(\ref{eq6}). Note that when the masking ratio is $0$, only the segmentation loss is used for training. Please zoom in for more details.}
      \label{fig:mask_ratio}
   \end{figure}

\begin{figure*}[t]
  \begin{minipage}{0.33\textwidth}
    \centering
    \includegraphics[width=\linewidth]{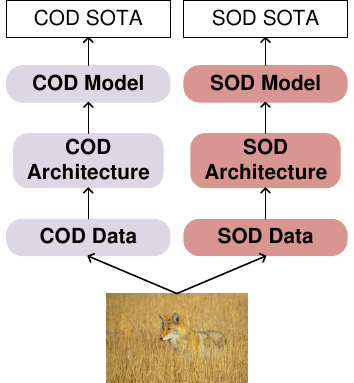}
    \subcaption{2 architectures, 2 models.}
    \label{fig:3_1}
  \end{minipage}
  \begin{minipage}{0.33\textwidth}
    \centering
    \includegraphics[width=\linewidth]{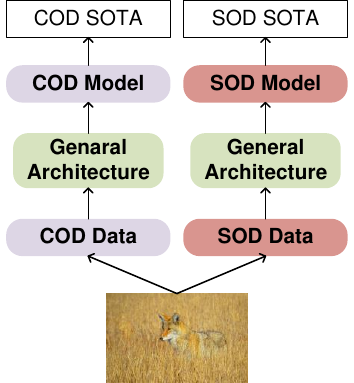}
    \subcaption{1 architecture, 2 models.}
    \label{fig:3_2}
  \end{minipage}
  \begin{minipage}{0.33\textwidth}
    \centering
    \includegraphics[width=\linewidth]{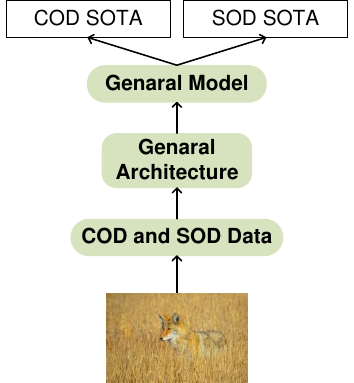}
    \subcaption{1 architecture, 1 model.}
    \label{fig:3_3}
  \end{minipage}
  \caption{\small (a) Previous methods developed specialized architectures and models for each task to achieve SOTA performance. (b) SENet (Ours) uses the same architecture to achieve SOTA performance across different tasks. However, we still need to train different models for different tasks, resulting in a semi-universal approach. (c) The two joint training paradigms we propose try to use one architecture to train one model that can achieve SOTA performance on both tasks at the same time.}
\label{fig:3_comparison}
\end{figure*}
\vspace{1mm}
\noindent \textbf{Qualitative analysis}. Fig.~\ref{fig:sod_results} presents a visual comparison of SENet with some representative competitors in several typical scenarios. It can be seen that when the target object is small, other methods \cite{f3net, MINet, LDF, UJSC, SelfReformer, TGSD, EVP} either fail to locate the target or make completely erroneous shape predictions. Only our method provides relatively accurate predictions. When the target object is larger and the prediction difficulty is lower, other methods can achieve approximately accurate predictions but exhibit roughness in some small details. Our method can achieve finer predictions in such cases.

\subsection{Masking Ratio}
\label{exp:mask}

As shown in Fig.~\ref{fig:mask_ratio}, we conduct experiments on the impact of using different masking ratios on network performance, and report the results on two COD datasets and two SOD datasets. It can be seen clearly that a smaller masking ratio (e.g., $5\%$) can improve performance on both tasks, which proves that using image reconstruction as an auxiliary task can indeed improve the performance of SENet. We speculate it may be because reconstruction requires the network to learn more low-level information such as texture, outline and edge, etc. \cite{maehog}, which is very helpful for segmentation. 

The difference is that COD is more sensitive to the masking ratio, only a smaller masking ratio can improve performance. As the masking ratio increases, the overall performance decreases obviously. Especially when the masking ratio exceeds $60\%$, the performance drops significantly. SOD is not sensitive to the masking ratio, and its performance has always been relatively stable. Only when the masking ratio exceeds $80\%$ can a significant decline be seen, and the decline is much smaller than that of COD.

A possible explanation is that the reconstruction difficulty of the camouflaged images is indeed high, the network is likely to reconstruct the concealed objects to the similar surroundings \cite{UJSC-V2}, while the reconstruction of the salient images is relatively easier. As demonstrated in \cite{MAE}, it can effectively reconstruct a normal image with a higher masking ratio (e.g., $75\%$). The images in COD are very special and different from normal images, but SOD is not.

As shown in Table~\ref{table:efficiency}, we report the calculation amount and training time corresponding to different masking ratios. It can be seen that as the masking ratio increases, the calculation efficiency can be continuously improved and the training time can be shortened. Combined with Figure 6 in the main manuscript, for COD, a $40\%$ masking ratio can achieve a good trade-off between performance and computing efficiency, sacrificing about $5\%$ of performance while gaining about $25\%$ of efficiency gain, for SOD, this ratio can reach $70\%$, which can achieve a computing efficiency gain of about $50\%$ without affecting performance. Note that in this work we do not mainly emphasize computational efficiency, to achieve the best performance, we uniformly use a masking ratio of $5\%$ during training.

\subsection{Joint Training and Cross-domain Evaluation}
\label{exp:joint}

To better discuss the similarities and differences between COD and SOD, we conduct experiments of joint training and cross-domain evaluation. 
Table~\ref{table:joint} summarizes the quantitative testing results of models obtained through $4$ different training methods on $9$ datasets. 

First of all, it can be seen that the results of the two cross-domain evaluation are very poor compared with the other three cases, which also proves that there is a contradiction \cite{UJSC} between the two tasks, the model trained on one task cannot work well on another. 

Secondly, from the results of Paradigm 1, it can be seen that the joint training of COD and SOD has a negative effect on COD task but has no negative impact on SOD task and even slightly improves the performance on some datasets. This is also consistent with the view pointed out in previous work \cite{UJSC-V2} that “simple samples of COD are hard samples for SOD and can help improve the performance of SOD task, but not vice versa”. 

Finally, from the results of Paradigm 2, it can be seen that through a simple design of sharing the encoder and using independent decoders structure, the overall performance can be improved. Not only can it slightly improve the performance on SOD task, but it can also significantly reduce the damage caused by joint training to COD task. As for why joint training is not helpful for COD task, we speculate this could be related to the size of the training sets of the two tasks. The training data of COD is much less than SOD, and during joint training, our straightforward training strategy (see the Appendix for the pseudocode) might lead to model overfitting on COD task.

\subsection{Ablation Study}
\label{exp:ablation}

\noindent \textbf{Effectiveness of LICM}. Comparing the first and third rows in Table~\ref{tab:ablation} and also comparing the second and last rows, it can be seen that in both cases of using and not using the DW loss, the addition of the LICM effectively enhances the network's performance. LICM has a greater potential for improving the performance of COD, for COD task is more challenging and has greater room for improvement.

\vspace{1mm}
\noindent \textbf{Effectiveness of DW loss}.
Comparing the first and second rows in Table~\ref{tab:ablation} and also comparing the third and last rows, it can be seen that in both cases of using and not using the LICM, the addition of the DW loss effectively enhances the network's performance. Similarly, the DW loss leads to a greater improvement in COD task. In addition, from the comparison between the fifth row and the last row of Table~\ref{tab:ablation}, it can be seen that DW loss performs better than another widely used weighted loss \cite{f3net}, which also proves the effectiveness of the proposed DW loss.

Moreover, from the comparison between the fourth and last rows in Table~\ref{tab:ablation}, it can be seen that the bilinear interpolation method can effectively improve the performance of the network as it addresses the issue of labeling uncertainty to a certain extent, thereby improving the quality of ground truth images. The improvement is more significant in COD for labeling uncertainty is more prevalent in COD datasets.

\vspace{1mm}
\noindent \textbf{Impact of different initialization}. From the comparison between the sixth and last rows in Table~\ref{tab:ablation}, it can be seen that initializing the encoder with weights obtained through supervised training \cite{ViT} leads to a significant decrease in network performance compared to image reconstruction pre-training \cite{MAE}. This further validates our assertion that image reconstruction task is beneficial for extracting more low-level information that aids COD and SOD tasks.

Furthermore, from the last two rows in Table~\ref{tab:ablation}, it can be seen that the role of the MAE pretrained decoder is crucial, and it is a key factor in achieving excellent performance for SENet. It is also hoped that this can inspire everyone to discover more about the effectiveness of the MAE decoder, rather than ignoring it.

\subsection{Analysis}
\label{exp:analysis}

We present a summary and discussion of distinctions between prior works and our work.  (a) Previous works have been accustomed to designing special architectures and training special models separately for COD and SOD tasks (see Fig.~\ref{fig:3_1}). This also results in that the architecture specifically designed for one task does not work for another, even though both are binary segmentation tasks, with almost no versatility. (b) Our SENet is a general architecture suitable for both COD and SOD tasks, but we need to train two models using different datasets (same architecture but different parameters) to complete these two tasks respectively (see Fig.~\ref{fig:3_2}), resulting in a semi-universal approach. (c) The two joint training paradigms (see Fig.~\ref{fig:joint_training}) we propose try to solve the problem of using one architecture to train one model to complete two tasks at the same time (see Fig.~\ref{fig:3_3}), which is truly universal. However, we have only preliminarily solved this problem and have not really achieved the best performance on both tasks. This is also what we want to solve in future work.

\section{Conclusion}
\label{sec:conclusion}
In this work, we propose the SENet, which achieves SOTA performance on both COD and SOD tasks. We point out that image reconstruction is a helpful auxiliary task for COD and SOD tasks. We introduce the LICM and DW loss, and also explore the issue of joint training of COD and SOD. Extensive experiments have verified the effectiveness of our proposed method. We note that the study of joint training of COD and SOD is still at an early stage. We will continue to explore it in the future, hoping to improve the performance of both tasks at the same time.


\ifCLASSOPTIONcaptionsoff
  \newpage
\fi

\bibliographystyle{IEEEtran}
\bibliography{IEEEabrv,reference}

\end{document}